\documentclass{article}
\usepackage{ijcai23}

\usepackage{times}
\usepackage{soul}
\usepackage{url}
\usepackage[hidelinks]{hyperref}
\usepackage[utf8]{inputenc}
\usepackage[small]{caption}
\usepackage{graphicx}
\usepackage{amsthm}
\usepackage{booktabs}
\usepackage{algorithm}
\usepackage{algorithmic}
\usepackage[switch]{lineno}

\usepackage{cite}
\usepackage{float}
\usepackage{amsmath,amssymb,bm}
\usepackage{amsfonts,amssymb}
\usepackage{newfloat}
\usepackage{listings}

\usepackage{natbib}  
\usepackage{diagbox}
\usepackage{color}
\usepackage{amsthm}

\title{A Distributed Computation Model Based on Federated Learning Integrates Heterogeneous models and Consortium Blockchain for Solving Time-Varying Problems}

\author{
    Zhihao Hao$^{1, 2, 3}$, Guancheng Wang$^1$, Chunwei Tian$^{1,4}$, Bob Zhang$^1$
    \affiliations
   $^1$ PAMI Research Group, Department of Computer and Information Science, University of Macau, Macau SAR, China\\
   $^2$ School of Data Science, Chinese University of Hong Kong, Shenzhen, Guangdong, China\\
   $^3$ Shenzhen Institute of Artificial Intelligence and Robotics for Society, Shenzhen, Guangdong, China\\
   $^4$ School of Software, Northwestern Polytechnical University, Xi’an, Shaanxi, China
    \emails
    hao.zhihao@connect.um.edu.mo, \{yc07455, bobzhang\}@um.edu.mo, chunweitian@nwpu.edu.cn
}

\begin{document}

\maketitle

\begin{abstract}
The recurrent neural network has been greatly developed for effectively solving time-varying problems corresponding to complex environments. However, limited by the way of centralized processing, the model performance is greatly affected by factors like the silos problems of the models and data in reality. Therefore, the emergence of distributed artificial intelligence such as federated learning (FL) makes it possible for the dynamic aggregation among models. However, the integration process of FL is still server-dependent, which may cause a great risk to the overall model. Also, it only allows collaboration between homogeneous models, and does not have a good solution for the interaction between heterogeneous models. Therefore, we propose a Distributed Computation Model (DCM) based on the consortium blockchain network to improve the credibility of the overall model and effective coordination among heterogeneous models. In addition, a Distributed Hierarchical Integration (DHI) algorithm is also designed for the global solution process. Within a group, permissioned nodes collect the local models' results from different permissionless nodes and then sends the aggregated results back to all the permissionless nodes to regularize the processing of the local models. After the iteration is completed, the secondary integration of the local results will be performed between permission nodes to obtain the global results. In the experiments, we verify the efficiency of DCM, where the results show that the proposed model outperforms many state-of-the-art models based on a federated learning framework.
\end{abstract}

\section{Introduction}

Utilizing the neural network to solve mathematical problems corresponding to complex environments in various fields has made great progress in recent years (\cite{kumar2011multilayer}). Especially for the widely existing time-varying problems, the recurrent neural network (RNN) model has been applied in many fields due to its outstanding performance in processing time series data (\cite{weerakody2021review,wang2023activated}). In particular, as a special RNN model designed to meet the challenges of time-varying problems, the zeroing neural network (ZNN) can observe the changing tendency of time-varying parameters by utilizing its time-derivative, which makes the ZNN model solve the time-varying problems without residual error (\cite{jin2017zeroing,wang2022new}). However, these methods also have certain limitations. For example, according to (\cite{yuan2022roadmap}), the performance of centralized ZNN models is greatly affected by many factors such as model scale, computing power, data and so on. Moreover, problems like isolation and lack of interactions from each other for centralized models can create model silos, which can lead to many challenges such as transparency, efficiency, credibility and so on (\cite{hao2023jas}). Therefore, centralized models cannot ensure the stability of their performance for problems caused by model and data silos that exist widely in reality. This has also become one of the important factors affecting the difficulty of implementing centralized artificial intelligence.

Here, distributed artificial intelligence (DAI) models can meet these challenges. The traditional DAI models have been proposed to meet growing scalability and computing demands. This is especially true for Federated Learning (FL), which is widely recognized as the next evolution of DAI (\cite{li2022blockchain, li2021survey}). In FL, data does not need to be shared. Each client uses local private data to feed the model separately, before sharing the parameters to the server, which can aggregate parameters for the model update. It is possible to accurately grasp the dynamic changes of the parameters in time-varying problems by cooperating to complete specific tasks. However, this method only realizes the distribution of homogeneous models without considering model heterogeneity, which is specifically manifested in the difference in performance due to the difference in models with different structure when processing various tasks. Also, heterogeneity models will exhibit dissimilar performance due to mismatched equipment. This creates an obstacle for its large-scale implementation in reality. Besides this, it still relies on a server for important operations, and the corruption of the server may cause the failure of the entire model.

In this case, decentralization is necessary, which can be implemented by the blockchain that can serve as a foundation system due to its feature of being fully-distributed. In addition, the characteristics of non-modifiable and traceability can establish trust between untrusted nodes (clients) in the network. However, issues caused by the consistency of the node authority in permissionless blockchain (e.g., public blockchain (\cite{ferdous2021survey})) can lead to inefficiencies. Thus, the division of the node authority in the network has become a feasible solution to improve efficiency, such as consortium blockchain (\cite{dib2018consortium}) and private blockchain (\cite{pahlajani2019survey}). In particular, the consortium blockchain ensures that the model is decentralized, while the division of the authority of the nodes can well correspond to the operations at different levels of the DAI model.

Motivated by these, we propose a Distributed Computation Model (DCM) based on a consortium blockchain network to process time-varying problems like dynamic linear equation in complex environments. In order to meet the challenges such as model heterogeneity (\cite{kairouz2021advances}), the proposed method uses various computation models to solve a problem, before integrating the calculation results collaboratively. Therefore, it does not require the aggregation process of the model parameters, since only the computation results are transferred among nodes. The process of each iteration guarantees the participation of all nodes through weight assignment to the results of each model. The aggregation process allows efficient communication across nodes with heterogeneous models, which enables effective collaboration based on the FL framework with heterogeneous models. Hinged on this, the main contributions of this paper can be summarized as follows:

(i) We propose a novel DCM to address the growing challenges faced by centralized models. Furthermore, the model solves the time-varying problems through collaboration between different nodes equipped with heterogeneous models.

(ii) To best our knowledge, this is the first time that a distributed hierarchical integration method has been proposed by using permissioned and permissionless nodes in the consortium blockchain, which can change the centralized local solution process to a distributed global solution process.

(iii) Extensive experiments demonstrate the superiority of the proposed method by outperforming many state-of-the-art models developed in a federated learning framework.

\section{Related Work}

The emergence of ZNN model provides the possibility to efficiently solve time-varying algebraic equations and optimization problems (\cite{chen2016improved, zhang2009global}). The traditional solution process is generally to achieve the minimum performance index by designing a recurrent neural network that evolves along the negative gradient descent. However, due to the variability of time-varying coefficients, any intrinsic method designed for computing static problems cannot guarantee the performance of time-varying problems, which may lead to task failure (\cite{jin2017zeroing}). Therefore, (\cite{zhang2002recurrent}) proposed a RNN model for solving the time-varying Sylvester equation to meet these challenges, which is considered to be the origin of ZNN. In recent years, this model has been improved by combining with many methods such as integral feedback control (\cite{jin2016noise}), Taylor's formula expansion (\cite{liao2015taylor}) and so on. Moreover, it has been used in many fields such as the motion planning of a robotic manipulator (\cite{wang2022convergence}), acoustic location system (\cite{huang2020modified}) and so on. However, the performance of these models is greatly affected by the sampling point and time interval, with the traditional centralized solution process causing the model to fail to obtain timely feedback from the global environment, and resulting in large errors eventually.

In order to solve the shortcomings of centralized learning, the DAI model came into being. In recent years, the distributed learning method represented by FL has been greatly developed in many fields such as finance (\cite{long2020federated}), intelligent medicine (\cite{polap2021agent}), face recognition (\cite{niu2022federated}) and so on. And there has been more and more work on the improvement of FL. For example, (\cite{wu2022smartidx}) proposed SmartIdx, a convolutional neural network (CNN)-based federated learning compression algorithm that reduces the overall communication cost by achieving a high compression ratio. In particular, for the current model and data heterogeneity problems faced by FL, (\cite{tan2022fedproto}) achieved a performance improvement by sharing prototypes, and also ensured the efficiency in its communication. However, the architecture of these models still rely on a server for aggregation. Hence, there is still no guarantee in the safety of the model and the global solution of the results.

Consortium blockchain provides a solution to these problems. Allocating different responsibilities according to the permissions of the nodes in the network to improve the overall performance has recently been applied in many fields. For example, after the user node completes the transaction, the information can be reviewed by the scheduling node before being uploaded to the blockchain (\cite{mao2019novel}) to ensure its credibility. Besides this, after the user node completes the transaction, the regulatory node collects reviews and invokes the deep learning model for sentiment analysis and prediction (\cite{hao2021novel}) to provide a reference for intelligent regulation. In addition, there are precedents for incorporating FL. For example, (\cite{chai2020hierarchical}) proposed a blockchain-driven hierarchical federated learning algorithm for knowledge sharing in the Internet of Vehicles, where it can resistant malicious attacks. In addition,  (\cite{hao2023novel}) designed an isomerism learning framework to enable efficient collaborations among heterogeneous models and extended to many fields such as sentiment analysis and public opinion management. All of these provide useful guidance for this work.

\section{Architecture of the Proposed System}

\begin{figure*}[htb]
\centering
\includegraphics[width=5in]{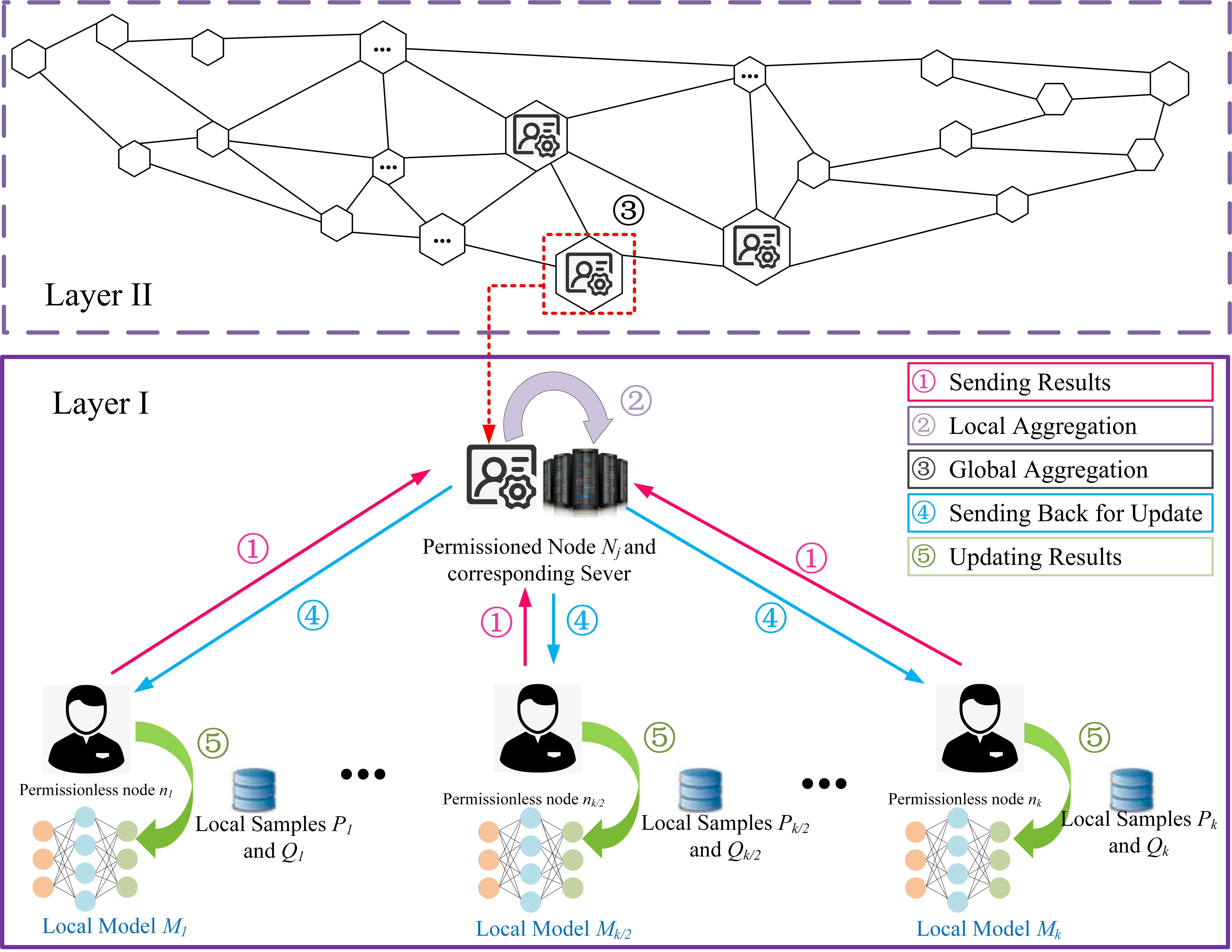}
\caption{Architecture for the proposed system.}
\label{fig_framework}
\end{figure*}

Since the nodes in the consortium blockchain network can be divided into permissioned and permissionless nodes according to the consensus algorithm. Therefore, in the process of dynamic aggregation of the distributed results, given two types of nodes can perform operations at different levels, the overall architecture can be divided into two layers (as shown in Figure \ref{fig_framework}). This hierarchical processing ensures that it incorporates the ideas of the master-worker architecture of FL (\cite{yang2019federated0}) based on the peer-to-peer architecture. The system can be divided in to different groups according to the number of permissioned nodes. Each permissioned nodes corresponds to different number of permissionless nodes. $I$ permissionless nodes with different computation models in a group can collaboratively solve the overall problem with the help of one permissioned node. Within a specific group, permissionless nodes are equipped with the same models, which are different from models in other groups. In particular, relevant information will be recorded in a shared ledger for all parties to view. Through this process, the system increases trust between anonymous nodes without the participation of a third-party. In addition, the security of the model has also been protected because the recorded information only contains the results after each iteration and does not expose the specific parameters of the local models. The process of the proposed system can be divided into five steps as follows:

\textbf{Step 1:} Permissionless nodes execute the solution process locally, and send the corresponding parameter to the permissioned node in their group.

\textbf{Step 2:} The permissioned node executes the operation of  aggregation after receiving all results from the permissionless nodes within the group.

\textbf{Step 3:} The permissioned node sends the local results to other permissioned nodes for aggregation to attain the global results.

\textbf{Step 4:} The permissioned node sends the aggregated results back to the permissionless nodes.

\textbf{Step 5:} Permissionless nodes update the local results for the next iteration process.

The whole iteration process will continue until reaching the maximum number. And all nodes in the network will share the parameter after execution of the global solution process.

In order to improve the security of the entire system, we also designed a verification mechanism to avoid security risks caused by malicious behaviors like nodes sending tampered results for aggregation. Since all results recorded on the blockchain are visible, and Digital Signature Algorithm (DSA) is used to protect the integrity of the related results during transmission. Therefore, the permissioned nodes can collect the results and remove results that deviate greatly from the result cluster. The senders can be located and marked through these deviations. After multiple rounds, if the number of times the marked node sending deviations accounts for a higher proportion of the overall number of sending results, it will be converted to a node that implements malicious behaviors, and the system can set it to a silent state, that is, it will no longer participate in the aggregation process from the next round of iteration.

Furthermore, there may be packet loss and other problems during the process of broadcasting results in Step 3, where noise will also affect the accuracy of the data to a certain extent. Therefore, in order to reduce the impact of the results being tampered with during transmission, we have also enhanced the consensus mechanism based on the Practical Byzantine Fault-Tolerant (PBFT) algorithm. Specifically, when the permissioned node completes the global aggregation operation, it will broadcast the corresponding result to others again. After a period of time, all permissioned nodes will count the results, and the result with the highest number of occurrences will be used for the next iteration. In this way, the error rate of the results can be reduced and the validity of the results can be improved.

\begin{figure}[!t]
	\centering
	\includegraphics[width=3.5in]{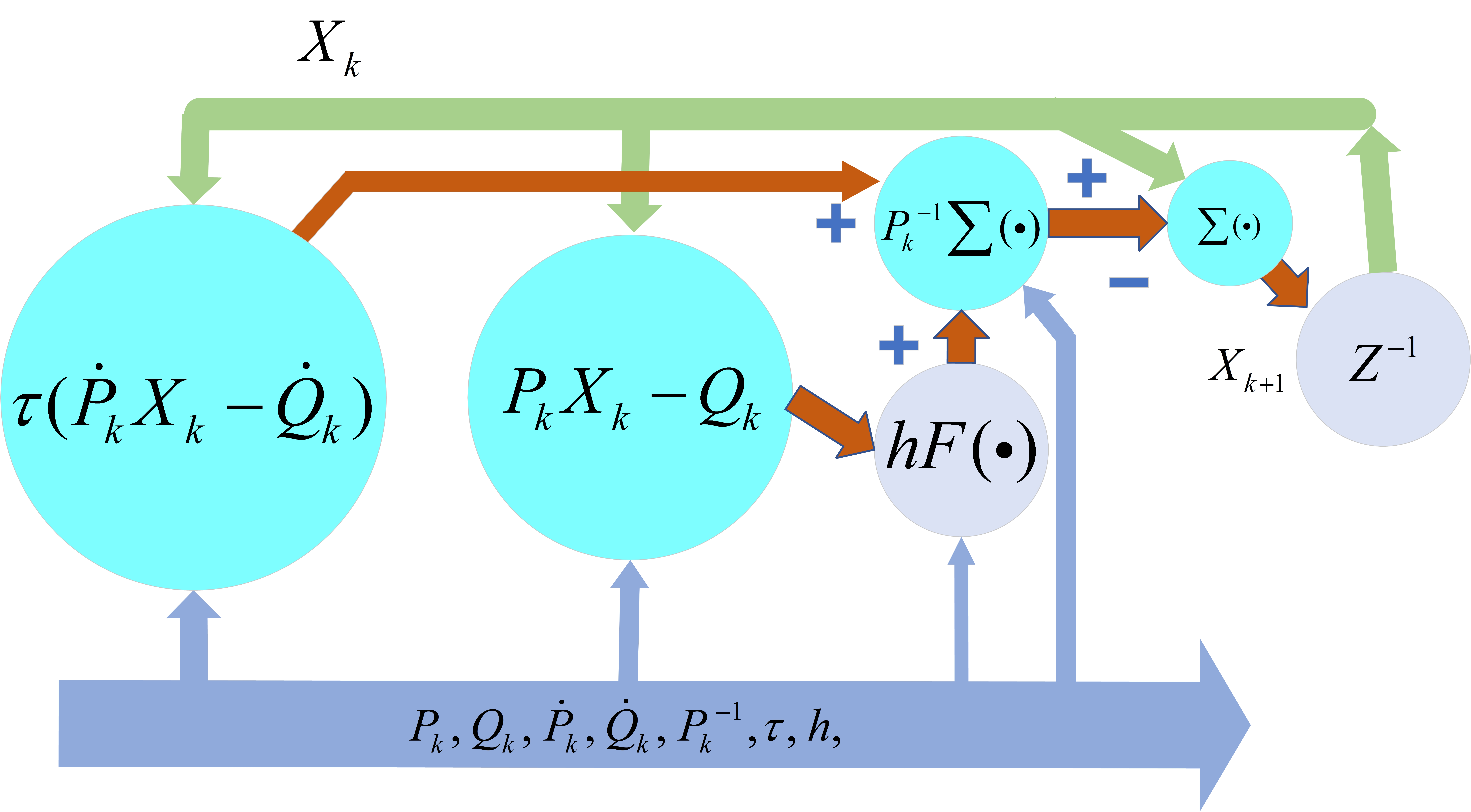}
	\caption{The ZNN model \eqref{ZNN-D} for solving the dynamic linear equation.}
	\label{fig_ZNN_model}
\end{figure}

\section{Heterogeneous Zeroing Neural Network (ZNN) Models}

In this paper, the DCM model is exploited to solve a time-varying linear equation, which can be expressed as

\begin{equation}\label{lin}
	P(t)X(t)= Q(t),
\end{equation}

where $X(t)\in \mathbb{R}^{n \times m}$ denotes the unknown matrix, $P(t) \in \mathbb{R}^{i \times n}$ and $Q(t) \in \mathbb{R}^{i \times m}$ are two time-varying parameter matrices. Among various permissioned nodes, various ZNN models are utilized to solve \eqref{lin}. The main idea of the ZNN model is to establish an error function as

\begin{equation}\label{error}
	G(t)=P(t)X(t)-Q(t),
\end{equation}

whose theoretical value is zero. Then, the ZNN model defines the time derivative of $G(t)$ as

\begin{equation}\label{ZNN-C}
	\dot{G}(t)=-\beta\digamma(G(t)),
\end{equation}

where $\gamma>0$ and $\digamma(\cdot)$ denotes an activation function. Based on \eqref{ZNN-C}, the ZNN model can make $G(t)$ converge to zero, where the corresponding $\xi(t)$ is the solution of \eqref{lin}. Inserting \eqref{error} into \eqref{ZNN-C}, the ZNN model can be formulated as

\begin{equation}\label{ZNN-C1}
	\dot{\xi}(t)=P^{-1}(t)(-\dot{P}(t)X(t)+\dot{Q}(t)-\gamma\digamma(P(t)X(t)-Q(t))),
\end{equation}

where $h=\beta\tau$ and $^{-1}$ denotes the inversion of a matrix (\cite{jin2017zeroing}). When implementing the ZNN model in digital circuit, \eqref{ZNN-C1} can be discretized at $t=k\tau$ as

\begin{equation}\label{ZNN-D}
	X_{k+1}=X_k-\theta_k,
\end{equation}

where $\theta_k=P^{-1}_k(-\tau\dot{P}_kX_k+\tau\dot{Q}_k-h\digamma(P_kX_k-Q_k))$ is the update step, $\tau$ denotes the sampling interval, and $h=\tau\beta$. According to \eqref{ZNN-D}, the topology diagram of the neural units and corresponding connection architecture of the model are shown in Figure \ref{fig_ZNN_model}.
Based on the uniform formula of the ZNN model \eqref{ZNN-D}, various ZNN models can be derived. For example, the conventional ZNN model (\cite{jin2018noise}) utilizing a linear activation function, as well as the first ZNN model termed as ZNN$_1$ model we used in DCM can be expressed as

\begin{equation}\label{ZNN-D1}
	X_{k+1}=X_k-\theta_{1,k},
\end{equation}

where $\theta_{1,k}=P^{-1}_k(-\tau\dot{P}_kX_k+\tau\dot{Q}_k-h(P_kX_k-Q_k))$. Besides this, the second ZNN model (\cite{wang2022convergence}) termed as ZNN$_2$ model estimates its $\theta_{2,k}$ with a bounded activation function, where the function can be expressed as

 \begin{equation}\label{S}
 	\digamma(G(t))=\left\{
 	\begin{array}{ll}
 		1,& \hbox{if $G(t)>1$,}\\
 		G(t),& \hbox{if $-1\leq G(t)\leq1$,}\\
 		-1,& \hbox{if
 			$G(t)<-1$}.
 	\end{array}
 	\right.
 \end{equation}

The third ZNN model, i.e., the ZNN$_3$ model in this paper is the Newton-Raphson iteration algorithm, which is a special ZNN model with $\dot{P}_k=0$ and $\dot{Q}_k=0$ (\cite{zhang2008zhang}). The ZNN$_3$ model estimates its update step as $\theta_{3,k}=-P^{-1}_k(P_k\xi_k-Q_k)$. To improve the accuracy and robustness of the ZNN$_3$ model, an integration implemented model (termed as ZNN$_4$) was proposed (\cite{wang2021noise}), whose $\theta_{4,k}=-P^{-1}_k(P_kX_k-Q_k+\sum\limits^k_{i=0}(P_iX_i-Q_i))$.

\begin{algorithm}[t]
\caption{The Distributed Hierarchical Integration (DHI)}
\label{alg_1}
\begin{algorithmic}[1] 
\STATE \textbf{Permissionless nodes} $n_i$ executes:
\STATE Initialize the local model $M_{i, (i=1,2,...,I)}$ provided by $N_j$ and randomly generate $X_1$
\FOR {each iteration $k$=1,2,...,$K$}
\STATE Samples $P_k$ and $Q_k$
\STATE Obtain the generation result $\theta_k^i$, $G^i_k$, and $X_{k+1}$
\ENDFOR
\STATE Sends ($\theta_k^i$,$G^i_k$) to the nearest permissioned node $N_j$
\\ \hspace*{\fill} \\
\STATE \textbf{Permissioned nodes} $N_j$ executes:
\FOR {each model computation round $r$=1,2,..}
\STATE $N_j$ determines the set $\delta_r$ of $I$ permissionless nodes
\FOR {$n_i \in \delta_{r}$ \textbf{in parallel}}
\STATE Steps 1-7
\ENDFOR
\STATE $N_j$ executes \eqref{eq_1}
\STATE $N_j$ executes \eqref{eq_2}
\FOR {j=1,2,...,J}
\STATE $N_j$ sends ($\theta_{j,k}(X_k)$, $G_{j,k}(X_k)$) to other permissioned nodes for aggregation as \eqref{eq_0}
\STATE $N_j$ broadcasts the aggregated results $\Theta_k(X_k)$ to all permissionless nodes
\ENDFOR
\ENDFOR
\end{algorithmic}
\end{algorithm}

\begin{figure*}[!t]
	\includegraphics[width=7in]{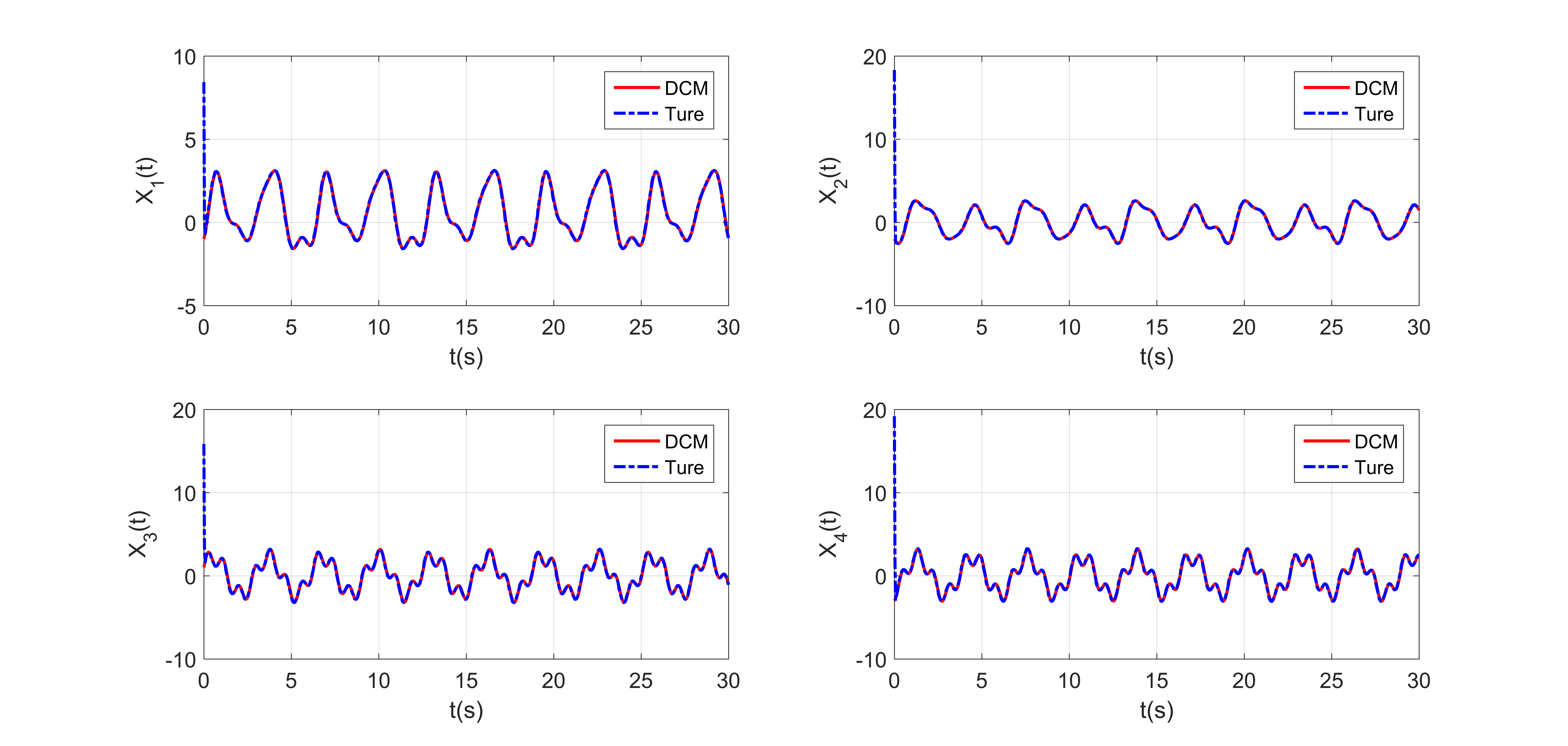}
	\caption{DCM compared with its true value.}
	\label{fig_compare_true_1}
\end{figure*}

\section{Distributed Hierarchical Integration (DHI)}

The proposed integrated scheme for collaborate iteration originated from the FedAvg algorithm (\cite{mcmahan2016federated}), which is used to meet the challenges existing in federated optimization such as non-iid data, unbalanced distribution of the data points, unstable communication links, etc. It has been verified to outperform some other federated optimizations like FedSGD (\cite{fekri2022distributed}). Influenced by this, we propose a Distributed Hierarchical Integration (DHI) algorithm based on the consortium blockchain network. DHI is applicable to the finite-sum of the objective function of different local models, and the objective function for $J$ models at the second layer are

\begin{equation}
\Theta_k(X_k) =\sum_{j=1}^{J}W_{j,k}\theta_{j,k}(X_k),
\label{eq_0}
\end{equation}

where $W_{j,k}=\frac{1/G_{j,k}(X_k)}{\sum\limits_{j=1}^J1/G_{j,k}(X_k)}$, $G_{j,k}(X_k)$ denotes the error of the $j$th group, and $\theta_{j,k}(X_k)$ can represent the update step of the $j$th permissioned node. Here, $W_{j,k}$ is used to maintain validity while guaranteeing decentralization, because it not only gives a large value to the models with good performance, but also keeps the involvement of the models with general effects. Besides this, $\theta_{j,k}(X_k)$ and $G_{j,k}(X_k)$ are estimated as

\begin{equation}
\theta_{j,k}(X_k)=\frac{1}{I}\sum\limits^I_{i=1} \theta^i_{j,k}(X_k)
\label{eq_1}
\end{equation} and

\begin{equation}
	G_{j,k}(X_k)=\frac{1}{I}\sum\limits^I_{i=1} G^i_{j,k}(X_k),
	\label{eq_2}
\end{equation}

respectively. Moreover, $\theta^i_{j,k}(X_k)$ represents the update step and $G_{j,k}(X_k)$ denotes the error of the $i$th permissionless node in a group belonging to the first layer.

DHI achieves model updating and results aggregation through hierarchical integration, where the model iterates the solution as
\begin{equation}
	X_{k+1}=X_{k}+\Theta_k(X_k) =\sum_{j=1}^{J}W_{j,k}\theta_{j,k}(X_k).
	\label{eq_3}
\end{equation}

It allows permissionless nodes to update their local models through multiple iterations and sends the results to the permissioned nodes for aggregation after each iteration process. After obtaining the local optimization results within different groups, the permissioned nodes will broadcast the results to each other for secondary aggregation and obtain the results of the global solution. The process of DHI is summarized in Algorithm \ref{alg_1}.

\section{Convergence Analysis}
To analyze the convergence of the DCM model, we provide the following theorem.

{\textbf{Theorem:} Utilizing the DCM to solve the time-varying linear equation, the computed solution globally converges to the theoretical one.

{\textbf{Proof:}
According to \eqref{eq_3}, we have

\begin{equation}
	X_{k+1}-X_{k}=\begin{bmatrix}W_{1,k}&W_{2,k}&...&W_{J,k} \end{bmatrix}\begin{bmatrix}\theta_{1,k}(X_k)\\\theta_{2,k}(X_k)\\...\\\theta_{J,k}(X_k) \end{bmatrix}
	\label{eq_4}.
\end{equation}

In addition, since $\sum_{j=1}^{J}W_{j,k}=1$, the left side of \eqref{eq_4} can be expressed as

\begin{equation}
	X_{k+1}-X_{k}=\begin{bmatrix}W_{1,k}&W_{2,k}&...&W_{J,k} \end{bmatrix}\begin{bmatrix}X_{k+1}-X_{k}\\X_{k+1}-X_{k}\\...\\X_{k+1}-X_{k} \end{bmatrix}
	\label{eq_5}.
\end{equation}

Combining \eqref{eq_4} and \eqref{eq_5} leads to

\begin{equation}
	\begin{bmatrix}W_{1,k}&W_{2,k}&...&W_{J,k} \end{bmatrix}\begin{bmatrix}X_{k+1}-X_{k}+\theta_{1,k}(X_k)\\X_{k+1}-X_{k}+\theta_{2,k}(X_k)\\...\\X_{k+1}-X_{k}+\theta_{J,k}(X_k) \end{bmatrix}=0
	\label{eq_6}.
\end{equation}

According to \eqref{eq_6}, the DCM can be deemed as a linear combination of various ZNN models. Since each ZNN model exploited in the DCM model globally converges to theoretical solution, the DCM also has global convergence. The proof is thus complete.

\begin{table*}[!t]
	\small
	\begin{tabular}{cllllllllll}
		\hline
		\diagbox{Models}{Time}        & 2.28             & 5.60            & 8.12             & 11.90            & 14.42           & 18.17            & 20.72            & 24.40            & 26.99            & 28.00            \\ \hline

		ZNN$_1$-FL        & 0.01853          & 0.01947         & 0.02009          & 0.02004          & 0.01893         & 0.02025          & 0.01910           & 0.02160           & 0.02083          & 0.02190           \\
		ZNN$_2$-FL     & 0.03618          & 0.03917         & 0.03965          & 0.04055          & 0.04013         & 0.04102          & 0.03952          & 0.04393          & 0.04115          & 0.04564          \\
		ZNN$_3$-FL        & 0.08272          & 0.14790          & 0.16150           & 0.13650           & 0.11430          & 0.14540           & 0.13930           & 0.19570           & 0.14780           & 0.27690           \\
		ZNN$_4$-FL        & 0.14960           & 0.15410          & 0.14850           & 0.15410           & 0.14870          & 0.15420           & 0.14870           & 0.15420           & 0.14850           & 0.14730           \\
		
		\textbf{DCM} & \textbf{0.01435} & \textbf{0.01150} & \textbf{0.01431} & \textbf{0.01165} & \textbf{0.01430} & \textbf{0.01185} & \textbf{0.01449} & \textbf{0.01175} & \textbf{0.01474} & \textbf{0.01941} \\ \hline
	\end{tabular}
	\caption{Comparison of errors of different models at selected time, which corresponds to specific rounds.}
	\label{tab:1}
\end{table*}

\section{Experimental Results and Analysis}
In this section, we evaluate the performance of the DCM. The experiment were run on Matlab 2020b with 128G memory, Intel(R) Core(TM) i3-3110M CPU and NVIDIA 1080Ti. Also, the test network consists of 4 $\times$ 5 nodes. There are a total of 4 permissioned nodes in the network, and each permissioned node corresponds to 5 permissionless nodes. Since the Ethereum platform provides relevant tools to build a consortium blockchain (\cite{zhang2018method}). It can be deployed on multiple virtual machines running Ubuntu Linux v16.04 in an Openstack environment. Each virtual machine is assigned 1 virtual CPU core, 2 GB of memory, and 10 GB of persistent storage. Furthermore, in order to reduce the impact of the communication process on the integration process, all virtual machines are connected together in a low-latency local network with an average round-trip time of less than 1 millisecond for communication between each node. The network behavior can be monitored by the Python Web3 Library and Elastic Search can be used for storage of the block information. Here, the sampling time interval for all models is set to 0.03s. And in order to reduce the influence caused by the computation time of different models, the communication time is set to 0.035s. Based on this, we ran the DCM model and observed the gap between the model's generation results and the true value in [0s, 30s].

In this experiment, the time-varying linear equation \eqref{lin} is defined as

\begin{equation}\label{parameter}
	\begin{aligned}
		&P(t)=\begin{bmatrix}
			\sin(2t) & ~~-\cos(2t)\\
			\cos(2t) & ~~\sin(2t)\\
		\end{bmatrix}\in \mathbb{R}^{2\times 2}
		,
		\\
		&Q(t)=\begin{bmatrix}
			(\sin(2t)+2) & ~~\cos(5t)+2\\
			-\cos(3t) & ~~\sin(5t)+1\\
		\end{bmatrix}\in \mathbb{R}^{2\times 2}
		.
	\end{aligned}
\end{equation}

The diagrams of the computed solution synthesized by DCM and the theoretical solution are shown in Figure \ref{fig_compare_true_1}. It can be seen from Figure \ref{fig_compare_true_1} that each element in $X(t)$ of the generated results of the DCM is very close to its true value, where the overall trend of the computed solution and true value are also fitted. According to Figure \ref{fig_compare_true_1}, the DCM can effectively solve the dynamic linear equation. To better demonstrate the effectiveness of the proposed DCM, we evaluated the steady-state error of the DCM and compared it with some state-of-the-art models including ZNN$_1$ model (\cite{jin2018noise}), ZNN$_2$ model (\cite{wang2022convergence}), ZNN$_3$ model (\cite{zhang2008zhang}), and ZNN$_4$ model (\cite{wang2021noise}). All of these models are modified based on the federated learning prototype (\cite{yang2019federated1}) to ensure effectiveness. For better presentation, all models ran for 857 rounds of integration, whose result is presented in Figure \ref{fig_compare_models}. In addition, we also found that the convergence speed of DCM is faster and the value will fluctuate within a certain range for a period of time, which is mainly due to the comprehensive effect of the performance of different integrated models. In particular, influenced by the oscillation of the integrated model (ZNN$_3$), the DCM model is steep and choppy.

\begin{figure}[!t]
\centering
\includegraphics[width=3in]{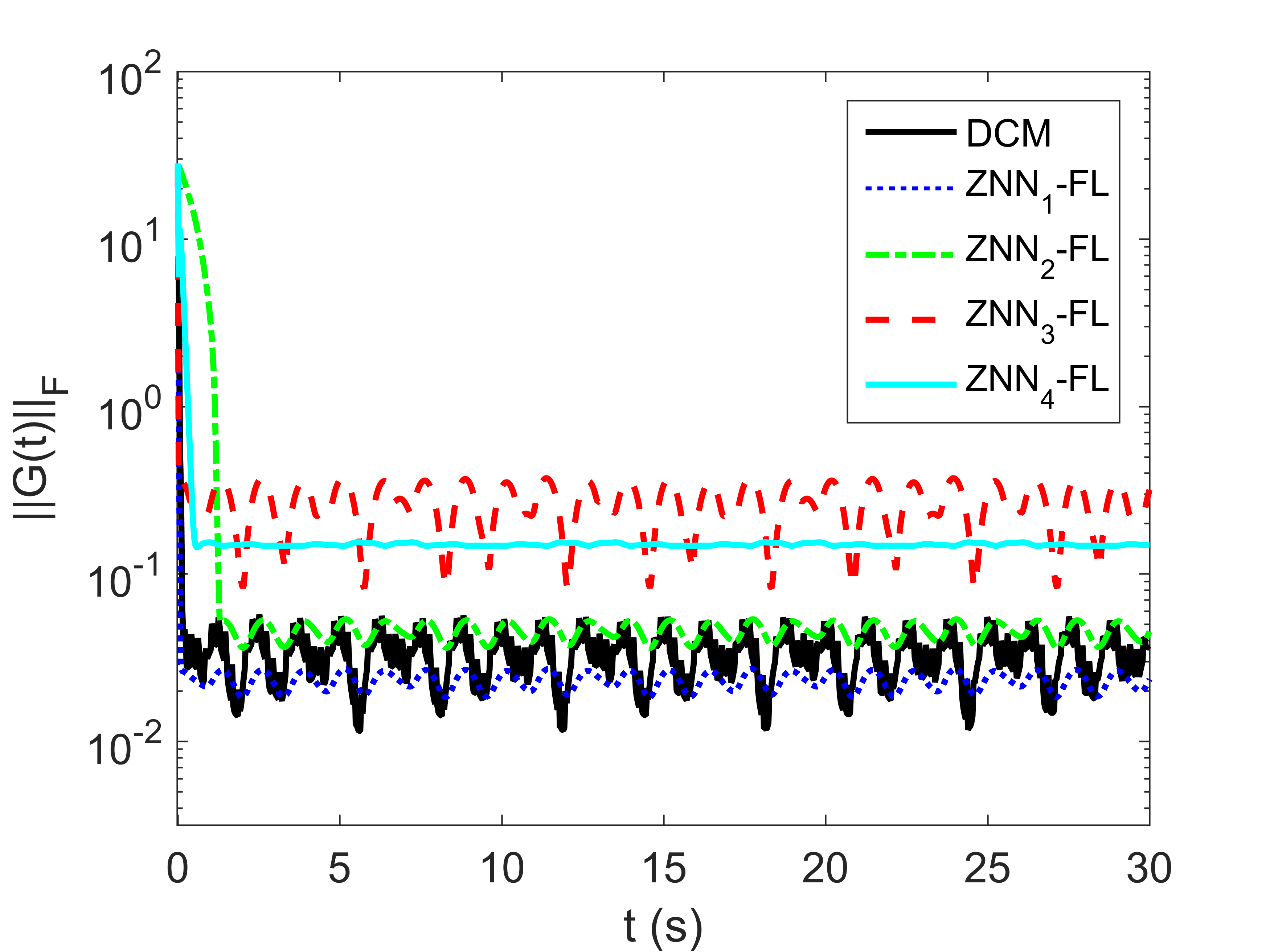}
\caption{Comparison of different models in federate learning ZNN$_1$-FL, ZNN$_2$-FL, ZNN$_3$-FL and ZNN$_4$-FL. Due to the imbalance of the communication time and sampling time, as well as the change of calculation results, the curve corresponding to each model is oscillating within a range.}
\label{fig_compare_models}
\end{figure}

It can be seen from Figure 4 that the proposed model is significantly better than the ZNN$_3$-FL model and the ZNN$_4$-FL model. However, it is difficult to see the difference between the proposed model and the ZNN$_1$-FL and ZNN$_2$-FL models. Therefore, for better analysis, we selected the errors of all models at 65, 160, 232, 340, 412, 519, 592, 697, 771 and 800 rounds for comparison, as shown in Table \ref{tab:1}. The average errors of the ZNN$_3$-FL model and ZNN$_4$-FL model are 15.4802 $\times$ $10^{-2}$ and 15.079 $\times$ $10^{-2}$, respectively. For the DCM model, the average value is 1.3835 $\times$ $10^{-2}$, which is the smallest compared to the average error of 2.0074$\times$ $10^{-2}$ for ZNN$_1$-FL and 4.0694 $\times$ $10^{-2}$ for ZNN$_2$-FL. Therefore, DCM is very efficient in terms of computational accuracy.

In addition, we also use different metrics to better evaluate the performance of these different models. Here, Mean Absolute Error (MAE) and Root Mean Square Error (RMSE) have been adapted because they can accurately measure the difference between the theoretical and calculated values. The value $E_{MAE}$ of MAE can be defined as $E_{MAE}=\frac{1}{n}\sum_{i=1}^{n}|true_{i}-computed_{i}|$, where $true_{i}$=0. And the value $E_{RMSE}$ of RMSE can be expressed as: $E_{RMSE}=\sqrt{\frac{1}{n}\sum_{i=1}^{n}(true_{i}-computed_{i})^2}$. Based on these, the relevant values can be shown in Table 2. As illustrated in this table, the DCM model method can outperform the many models like ZNN$_2$-FL, ZNN$_3$-FL and ZNN$_4$-FL because it has the lowest value in terms of $E_{MAE}$ and $E_{RMSE}$. Hence, the proposed method is effective. But there is still a slight gap compared to ZNN$_1$-FL (0.99 $\times$ $10^{-2}$ for MAE and 1.22 $\times$ $10^{-2}$ for RMSE). This is due to the weight distribution scheme adopted for decentralization. In the future, we will use higher-precision models for integration and try more weight distribution schemes to narrow the gap, such that the overall model performance is close to the higher-precision model. Furthermore, since centralized ZNN models have been widely employed in the control of redundant robots, the proposed DCM model can be exploited to solve the cooperative task execution problem in a distributed network among multiple redundant manipulators, where other distributed schemes have been designed and realized to maximize the manipulability or save the energy consumption (\cite{jin2020perturbed,li2016distributed,jin2017cooperative}).

\begin{table}[t]
\centering
\begin{tabular}{lll}
\hline
Models    & $E_{MAE}$    & $E_{RMSE}$   \\ \hline
ZNN$_1$-FL    & 2.30 $\times$ $10^{-2}$ & 2.31 $\times$ $10^{-2}$\\
ZNN$_2$-FL & 4.55 $\times$ $10^{-2}$ & 4.58 $\times$ $10^{-2}$ \\
ZNN$_3$-FL    & 26.29 $\times$ $10^{-2}$ & 27.47 $\times$ $10^{-2}$\\
ZNN$_4$-FL    & 14.93 $\times$ $10^{-2}$ & 14.93 $\times$ $10^{-2}$ \\
DCM  & 3.29 $\times$ $10^{-2}$ & 3.53 $\times$ $10^{-2}$ \\ \hline
\end{tabular}
\caption{Comparison between our method (DCM) and models using federate learning.}
\label{tab:2}
\end{table}

\section{Conclusion}
In this paper, we proposed a distributed intelligent computing model. The proposed model aims to solve the current model and data silos problems, achieving effective collaboration between different models. For the current model security and trust issues, we use a consortium blockchain network to achieve trusted interactions between unfamiliar nodes without exposing model parameters. Extensive experiments demonstrate that the proposed model performance is not only close to the high precision model, but also significantly outperforms most state-of-the-art models applying federated learning. To summarize, this work can provide a distributed scheme for the development of artificial intelligence driven by centralized frameworks such as the big model and generative artificial intelligence. Moreover, it also provides a potential solution for the construction of efficient swarm/crowd intelligence. Furthermore, it presents feasible opportunities for the implementation of trusted distributed general artificial intelligence. However, we find that the current work is still limited by some challenges such as smart contract vulnerabilities. In the future, we will consider effective solutions to these problems to improve the practicability of the overall system.

\section*{Acknowledgments}
This work was supported by Shenzhen Science and Technology Program under Grant ZDSYS2021102111141502 and the Shenzhen Institute of Artificial Intelligence and Robotics for Society.

\bibliographystyle{named}
\bibliography{ijcai23}

\end{document}